\documentclass[10pt,twocolumn,letterpaper]{article}

\usepackage[pagenumbers]{iccv} 

\definecolor{iccvblue}{rgb}{0.21,0.49,0.74}
\usepackage{multirow}
\usepackage{makecell}
\usepackage{dblfloatfix}
\usepackage{wrapfig}
\usepackage[pagebackref,breaklinks,colorlinks,allcolors=iccvblue]{hyperref}

\def\paperID{5} 
\def\confName{ICCV}
\def\confYear{2025}

\title{\textbf{\textit{ObjectTransforms}} for Uncertainty Quantification and Reduction in Vision-Based Perception for Autonomous Vehicles}

\author{%
\parbox{\textwidth}{\centering
\small 
\textbf{Nishad Sahu\textsuperscript{1}}\quad
\textbf{Shounak Sural\textsuperscript{1}}\quad
\textbf{Aditya Satish Patil\textsuperscript{2}}\quad
\textbf{Ragunathan (Raj) Rajkumar\textsuperscript{1}}\\[2pt]
$^{1}$Carnegie Mellon University, Pittsburgh, PA, USA\quad
$^{2}$University Of Minnesota, Twin Cities, MN, USA\\[2pt]
\texttt{nsahu@andrew.cmu.edu } \quad
\texttt{ssural@andrew.cmu.edu } \quad
\texttt{patil255@umn.edu} \quad
\texttt{rajkumar@andrew.cmu.edu }
}}

\begin{document}
\maketitle
\begin{abstract}
Reliable perception is fundamental for safety-critical decision-making in autonomous driving. Yet, vision-based object detector neural-networks remain vulnerable to uncertainty arising from issues such as data bias and distributional shifts. In this paper, we introduce \textit{\textbf{ObjectTransforms}}, a technique for quantifying and reducing uncertainty in vision-based object detection through object-specific transformations at both training and inference times. At training time, \textbf{\textit{ObjectTransforms}} perform color-space perturbations on individual objects, improving robustness to lighting and color variations. \textit{\textbf{ObjectTransforms}} also uses diffusion models to generate realistic, diverse pedestrian instances. At inference time, object perturbations are applied to detected objects and the variance of detection scores are used to quantify predictive uncertainty in real-time. This uncertainty signal is then used to filter out false positives and also recover false negatives, improving the overall precision–recall curve. Experiments with YOLOv8 on the NuImages 10K dataset demonstrate that our method yields notable accuracy improvements and uncertainty reduction across all object classes during training, while predicting desirably higher uncertainty values for false positives as compared to true positives during inference. Our results highlight the potential of \textit{\textbf{ObjectTransforms}} as a lightweight yet effective mechanism for reducing and quantifying uncertainty in vision-based perception during training and inference respectively.
\end{abstract}
\vspace{-6mm}
\section{Introduction}
\label{sec:intro}
\vspace{-1mm}
\begin{figure}
    \centering
    \includegraphics[width=1\linewidth]{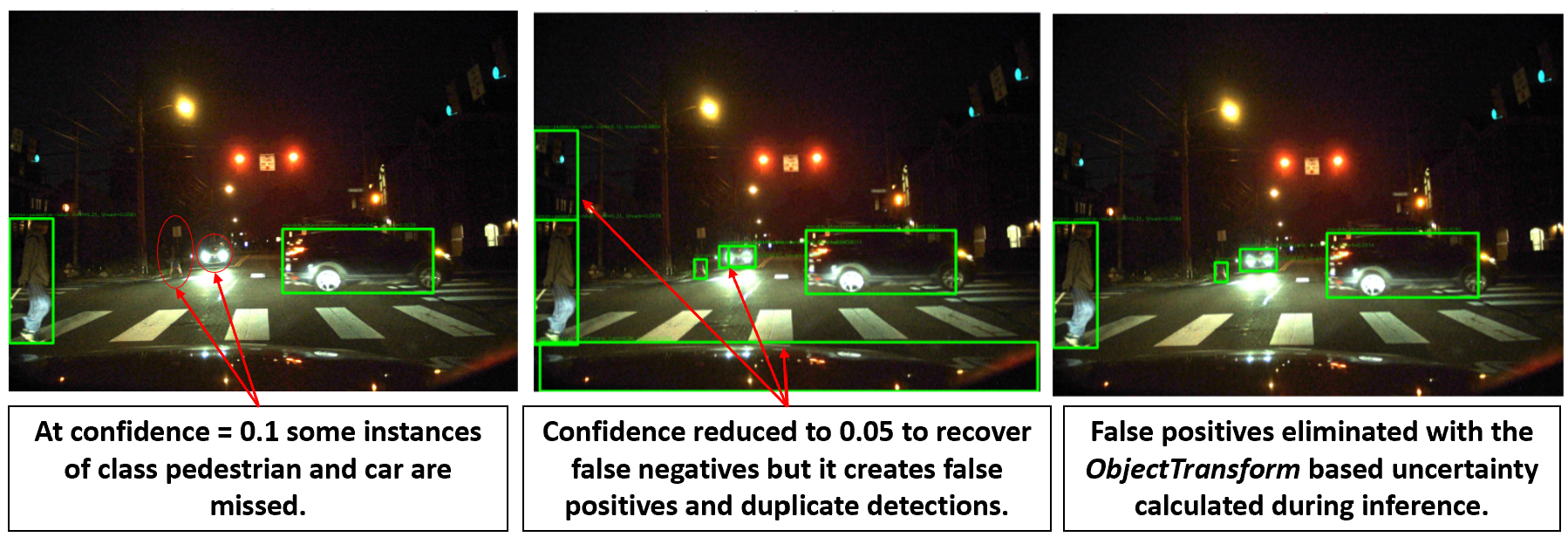}
    \caption{\textbf{Night-time crosswalk event captured from a real AV.} This is out of distribution data not used during training and validation of the YOLOv8 model. [Left] baseline detector at conf = 0.1 misses a pedestrian and a car. [Middle] Lowering to conf = 0.05 recovers these but introduces false positives. [Right] Our \textbf{\textit{ObjectTransforms}}-based uncertainty filtering suppresses the false positives while retaining the true positives.}
    \vspace{-8mm}
    \label{fig:dark}
\end{figure}

Autonomous vehicles (AVs) are poised to play an important role in increasing transportation safety, but reliable vision-based perception remains a major challenge for the wide-scale deployment of AVs due to the non-negligible occurrence of false positives and false negatives. False negatives, such as pedestrians being missed completely in low-light conditions, threaten the safety of all road users, while false positives can trigger very unsafe and/or unsettling maneuvers like phantom braking. Addressing these challenges requires methods to both quantify and reduce predictive uncertainty. A confidence score in a neural network represents how confident the network is about its output, whereas uncertainty indicates how reliable the network is in correctly detecting that output. Uncertainty arises due to noisy data, insufficient training data, an improper model and/or model weights \cite{kendall2017uncertainties}.  With valid uncertainty estimates, reliable sensor fusion can be done with other on-board sensors like lidars and radars, potentially augmented by V2X communication for safer downstream decision-making in AVs. 

\vspace{-0.1in}
In this paper, we propose \textbf{\textit{ObjectTransforms}}, a technique that applies object-specific augmentations at both training and inference stages to quantify and reduce uncertainty. At training time, \textbf{\textit{ObjectTransforms}} applies targeted object perturbations in the color space and diffusion-based pedestrian transformations to improve variability in the training dataset. At inference time, \textbf{\textit{ObjectTransforms}} performed controlled color perturbations to quantify predictive uncertainty, and enable the filtering of false positives and reducing false negatives.
\noindent Our contributions are threefold: (i) A novel theoretical formulation of uncertainty quantification as a violation of transformation invariance; (ii) Using \textbf{\textit{ObjectTransforms}} for increasing accuracy during training; and (iii) Using \textbf{\textit{ObjectTransforms}} at inference time to improve the overall area under the precision-recall curve.

\section{Related Work}
\label{sec:rw}
\vspace{-0.03in}
Neural networks can sometimes produce incorrect outputs with high confidence, reducing reliability and increasing uncertainty. The quantification of network uncertainty in itself is a challenge. Uncertainty quantification in neural networks is commonly performed using Bayesian methods such as Monte Carlo dropout and deep ensembles \citep{lyu2021uncertaintyestimationframeworkprobabilistic,lakshminarayanan2017deep,gal2016dropout}. Estimation of bounding box localization through Gaussian methods has also been explored \citep{ choi2019gaussianyolov3accuratefast}. Data augmentation is widely used to improve robustness in object detection. Techniques such as HSV jittering, CutMix \citep{yun2019cutmix}, and RandAugment \citep{cubuk2020randaugment} apply global transformations. However, global perturbations do not capture object-specific variability, leaving detectors sensitive to challenging scenarios like camouflaged pedestrians or lighting-induced appearance shifts. Recently, there has been a trend towards object-level augmentation in medical imaging and natural images \citep{zhang2021objectaug}, but the area is still underexplored in the context of autonomous driving. Diffusion models \citep{ho2020denoising} enable realistic data synthesis, offering opportunities for targeted augmentation. Test-Time Augmentation (TTA) \citep{shanmugam2021betteraggregationtesttimeaugmentation} provides uncertainty estimates, yet remains limited to image-level perturbations. Our work proposes object-specific transformations to quantify and reduce uncertainty. To the best of our knowledge, this is the first work exploring object-specific test-time augmentations in the context of AVs for quantifying and reducing uncertainty. 

\begin{table}[t]
\centering
\scriptsize
\caption{Notation used in the theoretical framework in section \ref{sec:theory}}
\vspace{-0.1in}
\begin{tabular}{ll}
\hline
\textbf{Symbol} & \textbf{Meaning} \\
\hline
$X$ & Input image \\
$o$ & Object instance in the image \\
$Y \in \{0,1\}$ & Ground-truth label: object present/absent \\
$\theta \in \Theta$ & Transformation parameters. \\
$q(\theta)$ & Sampling distribution of transformations \\
$T_\theta$ & Object-specific transformation with parameter $\theta$ \\
$o_\theta = T_\theta(o)$ & Object after applying transformation $T_\theta$ \\
$S(o_\theta) \in [0,1]$ & Detector confidence score for $o_\theta$ \\
$\tau$ & Detection threshold \\
$A_\tau$ & Detection event: $\{S(X_\theta) \geq \tau\}$ \\
$\mu$ & Ideal probability of $A_\tau$ under transformation invariance \\
$Z_\tau$ & Binary variable: $1$ if detection event $A_\tau$ occurs \\
$B_{\theta}$ & Event that a transformation with parameter $\theta$ is applied \\
$C$ & Scene context (geometry, pose, background, illumination) \\
$\mathbb{E}_{\theta}[\cdot]$ & Expectation over transformations $\theta$ drawn from $q(\theta)$ \\
$U(C)$ & Uncertainty score (in a given context C) \\
${U}_{class}(C)$ & Empirical variance of scores across transformations \\
$U_{\text{bbox}}(C)$ & Localization uncertainty (variance of box parameters) \\
\hline
\end{tabular}
\label{tab:notation}
\vspace{-0.2in}
\end{table}

\section{Uncertainty As a Violation of Transformation Invariance via \textbf{\textit{ObjectTransforms}}}
\label{sec:theory}
\vspace{-0.1in}
This section presents a theoretical framework for quantifying uncertainty in vision-based 2D object detection tasks using an invariance measure. Let $X$ denote an input image containing an object instance $o$ with ground-truth label $Y \in \{0,1\}$. We apply a object-specific transformation $T_\theta$ to the object $o$, where $\theta$ is sampled from a distribution $q(\theta)$. This operation of applying a transformation, denoted by $B_{\theta}$, produces a perturbed image $X_\theta$ which contains $o_\theta = T_\theta(o)$. A detector outputs a confidence score $S(o_\theta) \in [0,1]$, and for a threshold $\tau$ we define 
the \emph{detection event} $A_\tau := \{S(o_\theta) \geq \tau\}$. 
\vspace{-0.2in}
\paragraph{Transformation-Invariance Hypothesis.}
For a fixed scene Context $C$ (such as geometry, background, illumination outside the object mask), 
we postulate that the probability of detection should be independent of object-level transformations:
\begin{equation}
\footnotesize
    \Pr(A_\tau \mid B_{\theta}, C) \;=\; \Pr(A_\tau \mid C) \;=\; \mu , \quad \forall \theta \in \Theta.
    \label{eq:invariance}
\end{equation}
Equation~\eqref{eq:invariance} formalizes our intuition that reliable detectors must not rely on superficial changes in color, texture or appearance of a single object instance.

\vspace{-0.2in}
\paragraph{Uncertainty as a Violation of Transformation Invariance.}
To reason about invariance more clearly, let $Z_\tau$ be a binary random variable corresponding to the detection event $A_\tau$ that equals $1$ 
if the detector detects a transformed object $o_\theta$ in $X_{\theta}$ i.e. $S(o_{\theta}) \ge \tau$
and $0$ otherwise. 
If the detector is perfectly transformation invariant, $Z_\tau$ will not change across different transformations $\theta$, 
and its variance will be zero. By the \emph{law of total variance} \cite{wolter2007introduction}, the variability of $Z_\tau$ across transformations can be decomposed as
\begin{equation}
\scriptsize
\mathrm{Var}(Z_\tau \mid C) = 
\underbrace{\mathbb{E}_\theta[\mathrm{Var}(Z_\tau \mid B_{\theta}, C)]}_{\text{Noise}} 
+ \underbrace{\mathrm{Var}_\theta(\Pr(A_\tau \mid B_{\theta}, C))}_{\text{Effect of transformations}}.
\end{equation}

\noindent
The \emph{Noise} term  captures randomness internal to the detector 
(e.g., dropout or stochastic inference). The \emph{Effect of transformations} term measures how much the detection probability $\mu$ changes when we apply different transformations to the same object. If the detector is transformation invariant, probability is the same for all $\theta$, and the second term vanishes. Thus, variance across transformations acts as a direct measure of uncertainty. In other words, when predictions are transformation invariant, their variance across transformations ought to vanish. This motivates a practical definition of uncertainty as the variance. Detectors have both a classification confidence score as well as bounding box-coordinates for predictions. This motivates a practical definition of uncertainty as the variance of either classification scores or bounding-box coordinates:
\begin{equation}
\footnotesize
U_{\text{class}}(C) = \mathrm{Var}_\theta(S(o_\theta)), \qquad U_{\text{bbox}}(C) = \tfrac{1}{4}\sum_{d \in \{x,y,w,h\}} \mathrm{Var}_\theta(d).
\label{eq:variance_uncertainty}
\end{equation}

where, $U_{\text{class}}(C)$ captures the instability in classification confidence, while $U_{\text{bbox}}(C)$ captures instability in localization. 
We then combine them into a weighted sum.
\begin{equation}
\footnotesize
    U(C) = \omega_1 \, U_{\text{bbox}}(C) + \omega_2 \, U_{\text{class}}(C),
    \qquad \omega_1 + \omega_2 = 1,
    \label{eq:final_uncertainty}
\end{equation}
where the weights $\omega_1$ and $\omega_2$ can be tuned using a calibration or validation set.


\begin{figure}[t]
    \centering
    \includegraphics[width=\linewidth]{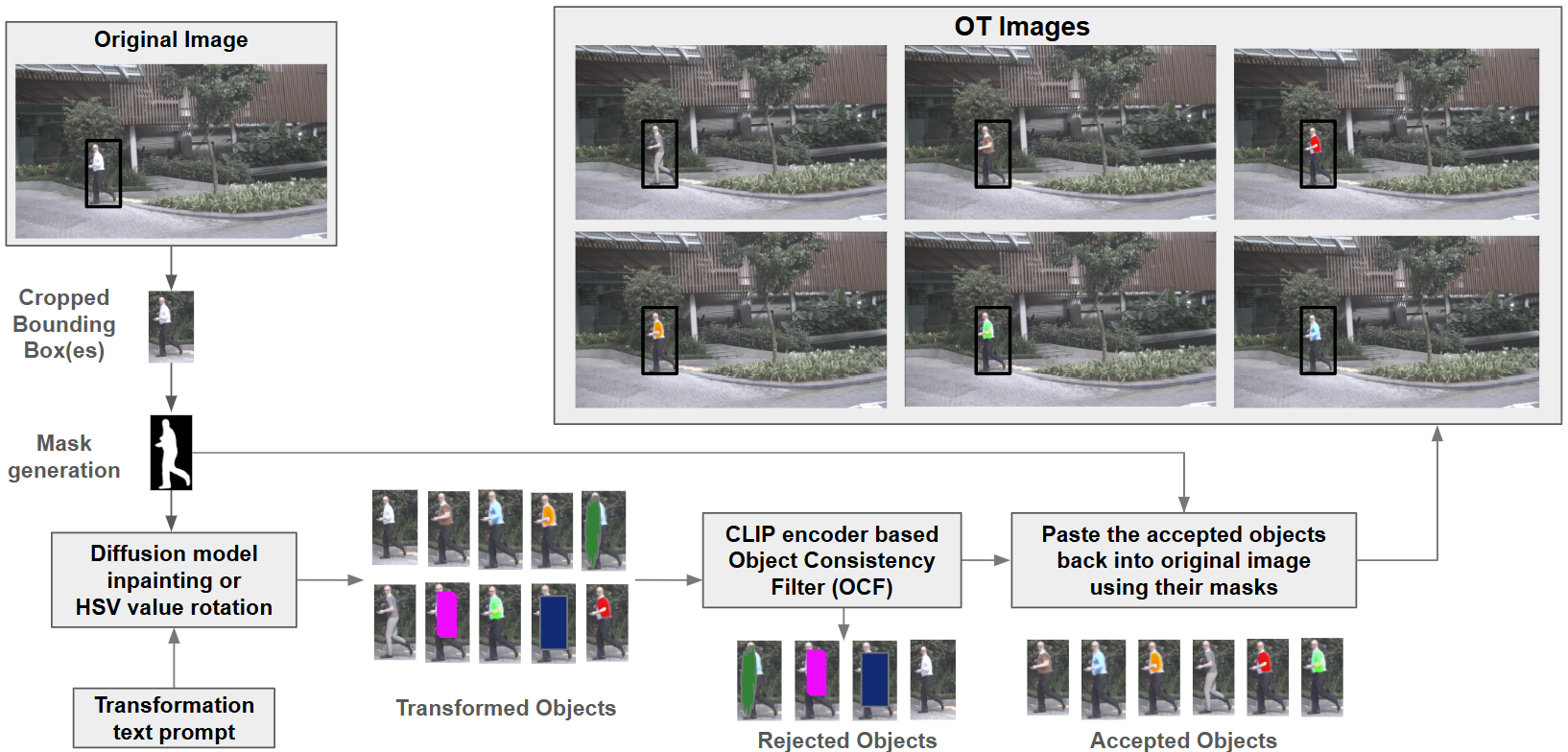}
    \caption{The diffusion-based pedestrian augmentation pipeline.}
    \label{fig:diffusion_model_pipeline}
  \vspace{-0.15in}
\end{figure}


\vspace{-0.2in}
\paragraph{\textbf{\textit{ObjectTransforms}}.}
In this paper, we instantiate $T_\theta$ as HSV perturbations and diffusion-based pedestrian augmentations, as concrete cases to our approach. In general, any object specific transformations such as object transforms in color space, addition of noise, crop, flip and rotations. can be applied. The unified uncertainty metric $U(C)$ thus quantifies overall invariance-based uncertainty. This allows us to filter false positives (high $U(C)$) while recovering stable low-confidence true positives.

\begin{table}[t]
\centering
\scriptsize
\caption{Detection Performance (mAP50-95) on NuImg10k}
\vspace{-0.1in}
\label{tab:augmentations}
\begin{tabular}{lccc}
\toprule
\textbf{Class} & \textbf{NuImg 10K} & \textbf{NuImg+ HSV (entire image)} & \textbf{\textit{ObjectTransforms}} \\
\midrule
Overall     & 0.384 & 0.383 & \textbf{0.407}\\
Pedestrians & 0.298 & 0.292 & \textbf{0.314}\\
Barriers    & 0.359 & 0.371 & \textbf{0.397}\\
Cones       & 0.35 & 0.344 & \textbf{0.382}\\
Vehicles    & 0.528 & 0.524 & \textbf{0.533}\\
\bottomrule
\end{tabular}
\label{tab:training}
\caption*{\textit{\textbf{Note}:} mAP50-95 is a strict metric resulting in values that might seem modest ($\approx$ 0.4). However, in practice, a lower value of IoU is typically used during deployment. For reference, YOLOv8 trained with \textbf{\textit{ObjectTransforms}} data achieves an mAP50 of over 0.6. While we use a 10K subset of nuImages for controlled experiments, full-scale training on the entire nuImages dataset ($>90K$ data points) is likely to push these metrics to much higher values.}
\vspace{-0.4in}
\end{table}

\vspace{-0.10in}
\section{Our Methodology}
\label{methodology}
\textbf{\textit{ObjectTransforms}} are applied during training time using two methods:  \\
\textbf{(1) Object-specific HSV transformations}: \textbf{\textit{ObjectTransforms}} instead apply object-specific HSV modifications: each mask of an object is randomly perturbed in hue, saturation or value and then reinserted at its original position into the image. For note: conventional HSV jittering applies global shifts across an entire image. This morphing significantly enriches object appearance variability while preserving the scene context. It also increases the range of robustness of the model across various illumination and camouflage conditions. HSV transformations are good for classes which are inanimate objects with well-defined shapes such as vehicles, barriers and cones.\\ \textbf{(2) Diffusion-based pedestrian augmentations}: HSV transformation in pedestrians may introduce change, to skin and hair colors which may not be realistic. In contrast, \textbf{\textit{ObjectTransforms}} takes a different approach to pedestrians by generating synthetic pedestrian samples with a diffusion model. The masks are first in-painted \citep{wasserman2025paint}. To maintain semantic consistency, an \textit{Object Consistency Filter (OCF)} uses CLIP embeddings such that only pedestrian images generated with high similarity to reference text embeddings are retained. Finally, images that are socially or ethically unacceptable are filtered out. Our diffusion model-based augmentation and filtering pipelines are illustrated in Figures \ref{fig:diffusion_model_pipeline} and \ref{fig:OCF_Filter} respectively. 

At inference time, \textbf{\textit{ObjectTransforms}} apply a controlled set of HSV perturbations to detected objects and rerun the detector. The variance in confidence scores across perturbations serves as an explicit uncertainty estimate: reliable detections maintain stable confidence scores, whereas those of ambiguous cases fluctuate considerably. This outcome enables filtering of unstable false positives that typically have higher uncertainty. Having the capability to filter false positives above an uncertainty threshold helps decrease the detection confidence threshold and yield fewer false negatives. To recover the false negatives we can decrease the confidence threshold. This may lead to increase in false positives which we can filter with the uncertainty threshold. In general, the area under the precision recall curves can improve notably with \textbf{\textit{ObjectTransforms}}.


\vspace{-0.10in}
\section{Experiments and Results}
We conduct two sets of experiments to evaluate the effectiveness of \textbf{\textit{ObjectTransforms}} in reducing uncertainty. The first assesses the effectiveness of HSV transformations and the second that of diffusion-based pedestrian augmentation. Across both sets, we use the \textit{Yolov8x} network for 2D object detection. Our baseline dataset is the \textit{nuImages} 10K dataset which contains 6,999 training, 1,515 validation and 1,484 test images. We maintain this partition in our experiments.\\

\begin{figure}[t]
    \centering
    \includegraphics[width=\linewidth]{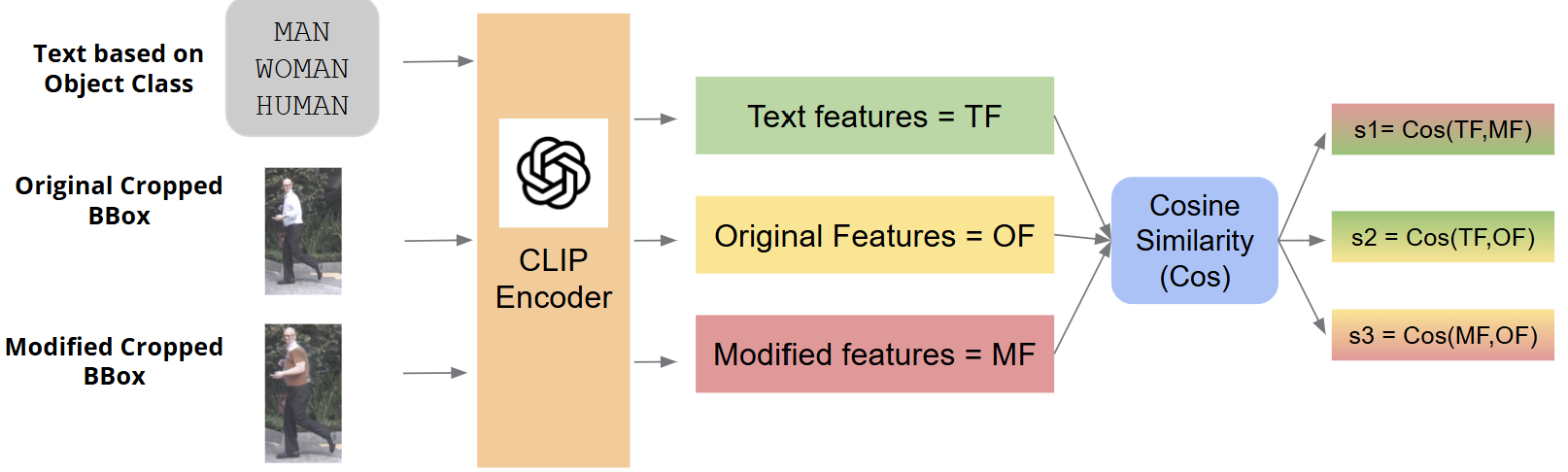}
    \caption{The Object Consistency Filter (OCF) using a CLIP encoder filters out incorrect outputs from diffusion-model inpainting}
    \label{fig:OCF_Filter}
  \vspace{-0.25in}
\end{figure}
\vspace{-0.15in}
\textbf{Experiment Setup 1}: We detect four classes:- pedestrians, vehicles, barriers and cones. We generate 97778 images (about 14 transformations of each image in the base dataset) using different object randomized HSV transformations across different object classes and their object instances in each training image. The \textit{Yolov8x} network is trained on (a) the base dataset, (b) the base dataset with image-level HSV augmentations and (c) the \textbf{\textit{ObjectTransforms}} dataset, each for 100 epochs. As shown in Table \ref{tab:training}, the \textbf{\textit{ObjectTransforms}} dataset results in significant gains in mAP50-95 scores relative to the base dataset and the base dataset with image-level HSV augmentations.
\begin{figure}{}
    \centering
    \includegraphics[width=\linewidth]{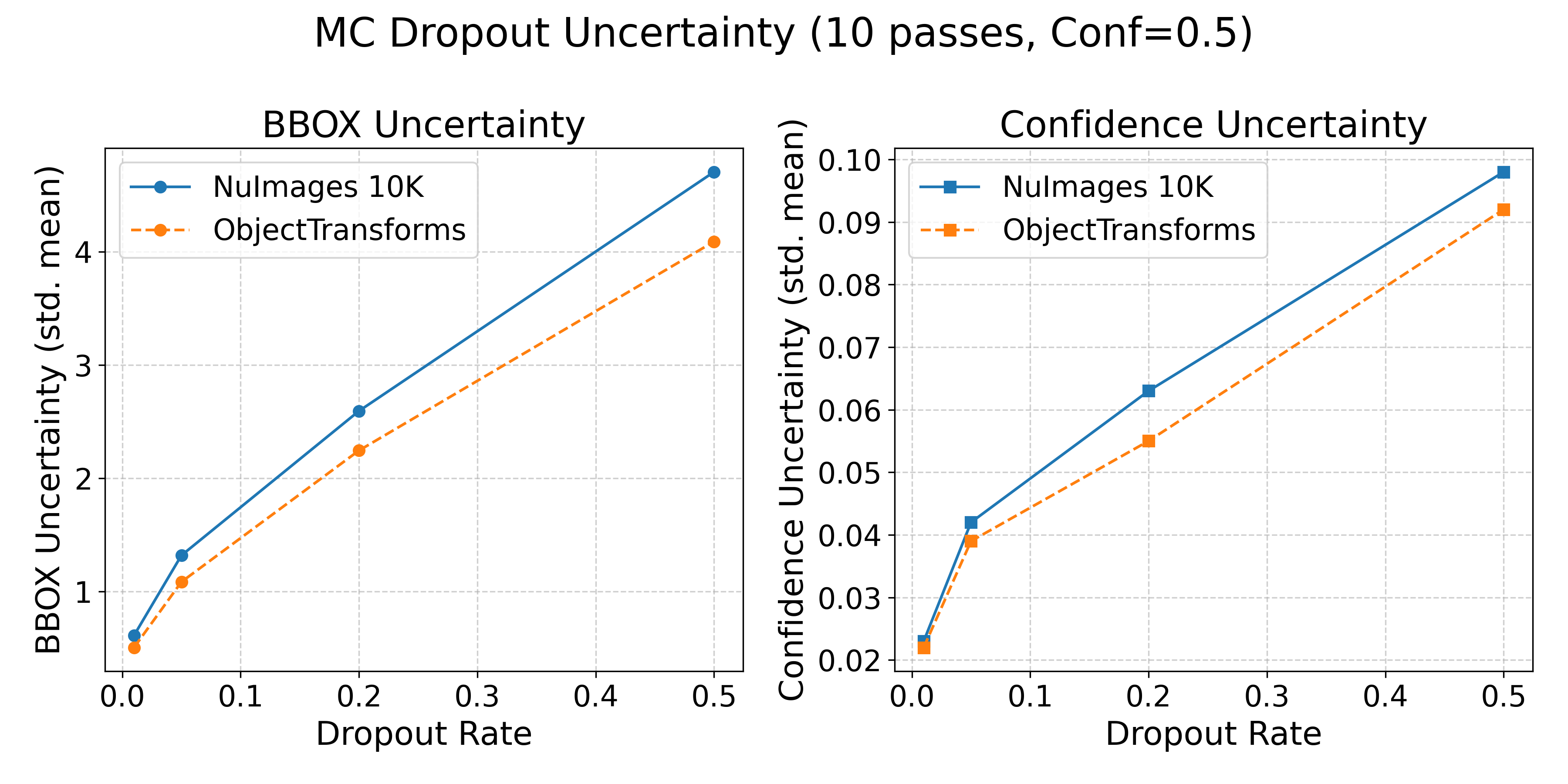}
    \vspace{-0.3in}
    \caption{MC Dropout Uncertainty on NuImages \citep{nuscenes} Test Dataset (10 passes, Conf=0.5).}
    \label{fig:mcdropout_uncertainty}
    \vspace{-0.15in}
\end{figure}

\begin{table}[t]
\centering
\scriptsize
\setlength{\tabcolsep}{2.5pt} 
\caption{Comparison of TP and FP with and without \textbf{\textit{ObjectTransforms}} (OT) during inference.}
\label{tab:tp_fp}
\begin{tabular}{lccccc}
\toprule
& \multicolumn{2}{c}{\textbf{Conf.=0.25}} & \multicolumn{3}{c}{\textbf{Conf.=0.01}} \\
\cmidrule(lr){2-3} \cmidrule(lr){4-6}
\textbf{Metric} & Without OT & \multicolumn{1}{c}{With OT} & Without OT & \multicolumn{2}{c}{With OT} \\
\cmidrule(lr){3-3} \cmidrule(lr){5-6}
& & $U_{th}{=}0.146$ & & $U_{th}{=}0.146$ & $U_{th}{=}0.19$ \\
\midrule
\textbf{TP}     & 3349 & 3186 & 4818 & 3719 & 4156 \\
\textbf{FP}     &  938 &  640 & 1864 &  685 &  908 \\
\textbf{TP/FP}  & 3.57 & 4.98 & 2.59 & \textbf{5.43} & 4.58 \\
\bottomrule
\end{tabular}
\end{table}

\begin{table}[t]
\centering
\scriptsize
\caption{Mean uncertainty scores for True Positives (TP) and False Positives (FP) obtained using the \textbf{\textit{ObjectTransforms}} inference-stage uncertainty quantification on the test set.}
\label{tab:uncertainty_tp_fp}
\vspace{-0.1in}
\begin{tabular}{lccc}
\toprule
\textbf{Metric} & \textbf{TP Mean} & \textbf{FP Mean} & \textbf{Separation (FP/TP)} \\
\midrule
$x$-uncertainty      & $4.02 \times 10^{-6}$ & $2.34 \times 10^{-5}$ & $5.82$ \\
$y$-uncertainty      & $2.98 \times 10^{-6}$ & $2.74 \times 10^{-5}$ & $9.20$ \\
$w$-uncertainty      & $7.25 \times 10^{-6}$ & $6.75 \times 10^{-5}$ & $9.31$ \\
$h$-uncertainty      & $9.44 \times 10^{-6}$ & $1.07 \times 10^{-4}$ & $11.36$ \\
Conf. ~uncertainty    & $6.26 \times 10^{-3}$ & $2.60 \times 10^{-2}$ & $4.16$ \\
\bottomrule
\end{tabular}

\vspace{-0.2in}
\end{table}
Next, we evaluate the inference-time uncertainty \textbf{U} across all detections using Monte Carlo dropouts for the model trained on the base dataset and the \textbf{\textit{ObjectTransforms}} dataset. 
where \textit{x,y} represent the center of a bounding box, \textit{h,w} are its height and width respectively and Var(.) is the statistical variance. Figure \ref{fig:mcdropout_uncertainty} summarizes the results: the model trained with \textbf{\textit{ObjectTransforms}} yields lower bounding box uncertainty (U\textsubscript{bbox}(C)) and confidence uncertainty (U\textsubscript{S}(C)). There is a consistent reduction in uncertainty of up to 20\% and the relative performance improvement increases with higher dropout rates. 

We next perform uncertainty quantification using \textbf{\textit{ObjectTransforms}} at inference time on our 1484 test images. The results are summarized in Table \ref{tab:uncertainty_tp_fp}. The uncertainty scores across all the parameters \textit{x, y, w, h} and confidence are much lower for the true positives compared to those of the false positives. Such substantive separation between the uncertainty of TPs and FPs helps in distinguishing between them, and hence isolate and highlight the FPs. For our test dataset, using the grid search technique, we found 0.25 and 0.75 to be good values of \textit{$\omega_1$} and \textit{\(\omega_2\)} respectively. With these values and a threshold of U\textsubscript{th} = 0.146, the framework preserves 95\% TPs and eliminates about 32\% FPs at a detection confidence threshold of 0.25. To recover false negatives, as shown in Table \ref{tab:tp_fp}, when we reduce the confidence threshold to 0.01 and U\textsubscript{th}=0.146 as before, we get a much higher TP with a decrease in FP as compared to Conf=0.25 with no \textbf{\textit{ObjectTransforms}}. 
\begin{figure}{}
    \centering
    \includegraphics[width=0.9\linewidth]{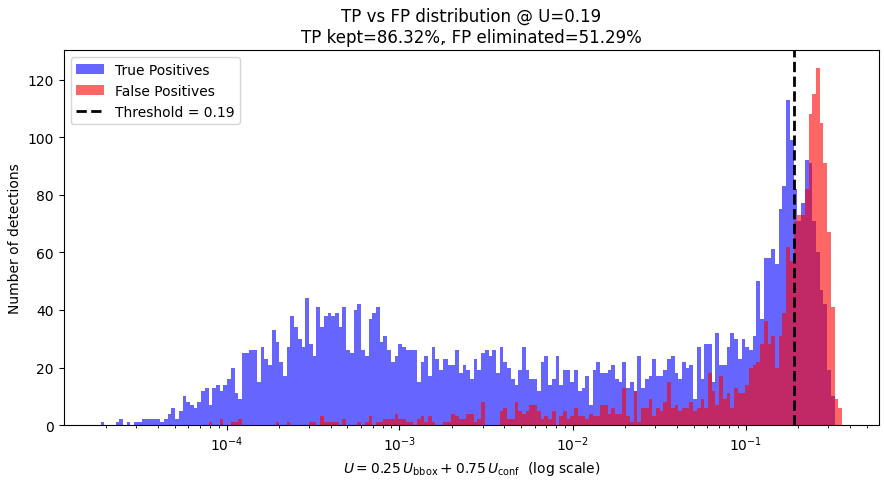}
    \vspace{-0.1in}
    \caption{Distribution of TP and FP along the uncertainty score. A threshold of U = 0.19 for conf = 0.01 (refer to Table \ref{tab:tp_fp})}
    \label{fig:tp_vs_fp_dist}
    \vspace{-0.25in}
\end{figure}
Furthermore, using U\textsubscript{th} = 0.19 results in a more significant increase in TP and a slight reduction in FP. These results are presented in Table \ref{tab:tp_fp} and are also graphically illustrated in Figure \ref{fig:tp_vs_fp_dist}. In practice, to find values of \textit{\(\omega_1\)}, \textit{\(\omega_2\)} and \textit{U\textsubscript{th}}, we can use a calibration set (similar to conformal learning \citep{gammerman2013learning}). 

\textbf{Real-time Feasibility}: With inference time uncertainty calculation using \textbf{\textit{ObjectTransforms}} we get 5 fps with the Yolov8 Extra Large model with a GPU usage of 2.5GB on a Nvidia L4 GPU. We expect to significantly improve the frame rate with lighter versions like \textit{nano}.

\textbf{Experiment Setup 2}: We detect only one class: pedestrians. The \textbf{\textit{ObjectTransforms}} dataset is generated using diffusion-based-inpainting \citep{wasserman2025paint} on the pedestrian instances in the base dataset and the OCF shown in Figure \ref{fig:OCF_Filter}. The size of the \textbf{\textit{ObjectTransforms}} dataset is 2072 images. We further performed a comparable analysis for this step and obtained similar results. Page length considerations limit an extended discussion.  


\vspace{-0.1in}
\section{Concluding Remarks}
\label{sec:conclusion}
We introduced \textbf{\textit{ObjectTransforms}}, a technique that applies object-specific transformations in both training and inference stages to reduce and quantify vision-based uncertainty, particularly for use in autonomous vehicle perception. The approach was evaluated using YOLOv8 on the nuImages 10K dataset. By performing color-wheel perturbations and diffusion-based pedestrian transformations, \textbf{\textit{ObjectTransforms}} improves mAP50-95 during training. Inference-time uncertainty estimates further enable significant improvements in filtering of false positives and recovery of false negatives. Together, these contributions highlight the promise of object-level transforms as a lightweight yet effective approach to quantify and reduce uncertainty for safer vision-based perception. In the future, we plan to study how object detection in low-visibility conditions can be improved.

{
    \small
    \bibliographystyle{ieeenat_fullname}
    \bibliography{main}
}

\end{document}